\documentclass[conf]{new-aiaa}
\usepackage[utf8]{inputenc}

\usepackage{soul}
\usepackage{algorithmic}
\usepackage{graphicx}
\usepackage{textcomp}
\usepackage{xcolor}
\usepackage{textcomp}
\usepackage[utf8]{inputenc}
\usepackage{amsmath}
\usepackage{float}
\usepackage[version=4]{mhchem}
\usepackage{siunitx}
\usepackage{longtable,tabularx}
\usepackage{algorithm}
\usepackage{lineno}
\usepackage{subfigure}
\usepackage{color,soul}
\title{Small jet engine reservoir computing digital twin}

\author{Colton J. Wright \footnote{Graduate Research Assistant, Mechanical Engineering, Ohio University, Athens, Ohio, 45701, Student Member}, Nicholas Biederman \footnote{Software Engineer, ResCon Technologies, LLC, 1275 Kinnear Road Suite 239, Columbus, Ohio, 43212}, 
Brian Gyovai\footnote{Founder/CEO, ResCon Technologies, LLC, 1275 Kinnear Road Suite 239, Columbus, Ohio, 43212}, Daniel J. Gauthier\footnote{Co-Founder, ResCon Technologies, LLC, 1275 Kinnear Road Suite 239, Columbus, Ohio, 43212}, and Jay P. Wilhelm\footnote{Associate Professor, Mechanical Engineering, Ohio University, Athens, Ohio, 45701, Senior Member}}

\begin{document}

\maketitle

\begin{abstract}
Machine learning was applied to create a digital twin of a numerical simulation of a single-scroll jet engine. A similar model based on the insights gained from this numerical study was used to create a digital twin of a JetCat P100-RX jet engine using only experimental data. Engine data was collected from a custom sensor system measuring parameters such as thrust, exhaust gas temperature, shaft speed, weather conditions, etc. Data was gathered while the engine was placed under different test conditions by controlling shaft speed.  The machine learning model was generated (trained) using a next-generation reservoir computer, a best-in-class machine learning algorithm for dynamical systems. Once the model was trained, it was used to predict behaviour it had never seen with an accuracy of better than 1.8\% when compared to the testing data. 
\end{abstract}

\section{Introduction}

The concept of a digital twin has gained attention in recent years as a means to simulate a complex physical system inside of a virtual environment. Reservoir computing is a lightweight machine learning (ML) algorithm that is well-suited for learning dynamical systems using time-series sensor data. The objective of this study was to explore the potential of a next generation reservoir computer (NG-RC) \cite{gauthier_next_2021} to create a jet engine digital twin.

An initial  ML model was created using numerically simulated jet engine data.  The model was able to accurately predict the engine thrust using an NG-RC model that included data from the requested and actual motor speed, exhaust gas temperature (EGT), and fuel-to-air ratio.  Based on the lessons learned from this study, an ML-based digital twin of a JetCat P100-RX micro-turbine engine was created.  The P100 model includes measurements of the turbine skin temperature.  The engine heats up during an extended engine run and this made the engine more efficient; this behavior is captured by the model by including the auxiliary temperature data.  

An initial numerical study was needed to demonstrate that the NG-RC is appropriate for jet engine modeling before moving on to experimental data. In the numerical study, the open-source Toolbox for Modeling and Analysis of Thermodynamic Systems (T-MATS) maintained by the National Aeronautics and Space Administration (NASA) was used \cite{chapman_propulsion_2014}.  The T-MATS default single-scroll gas turbofan engine model with a proportional-integral speed controller was used in the dynamical simulator to predict the behavior of the engine.  Step changes in the requested shaft speed were made, starting at nearly full throttle, stepping down to close to idle speed, then stepping up to full throttle.  This notional flight profile was similar to that used in the experiments.

A JetCat P100-RX micro-turbine engine, which weighs 1 kg and provides 100 N of thrust was used to gather experimental data for model training and validation. The engine has built-in sensors for several parameters, including shaft speed and EGT, which are important variables for predicting thrust.  Additional measurements of the engine temperature were collected by affixing thermocouples to the skin with high-temperature tape, which were read-out using a high-resolution analog-to-digital converter.  Ambient weather conditions were also measured once per day of engine testing so that the NG-RC's sensitivity to weather conditions could be accounted for. Once the experimental data was collected, data from engine tests were split into validation data and training data. Training data was used to create the NG-RC model, and validation data was used to evaluate the performance of the NG-RC prediction. The viability of using a NG-RC to create a jet engine digital twin was evaluated in this work.

\section{Literature Review}

The concept of a digital twin was first explored by Michael Grieves in 2002 \cite{kahlen_digital_2017}. The digital twin concept was originally created to assist with product life-cycle management. The primary goal of a digital twin is to model as much of a system as possible to replace some of the time, energy, and material costs of the design process with information. Information comes with the added bonus of revealing Unpredicted Undesirable (UU) behavior inside of the system. Grieves originally envisioned that the digital twin would fully describe the complex physical system, from the atomic level to the large scale geometrical level \cite{kahlen_digital_2017}.

In the optimal digital twin, any information that could be gained by inspecting the physical object could also be gained from the twin. One primary characteristic of the digital twin is the connection between the physical system and the twin. Sensors on the physical object feed information back to the twin to update the model for more accurate predictions. NASA was heavily invested in the digital twin concept throughout the 2010's because it would allow them to reduce prototyping costs and learn what UU behavior is present in their complex systems. NASA stated that the optimal digital twin has a technology readiness level of 1, but constituent capabilities are under development as of 2015 \cite{noauthor_nasa_2015}.

Capabilities to model an entire system in the way Grieves envisioned is not currently possible, but models can be made that predict the behavior of a complex system. Currently, the majority of digital twin applications are manufacturing related \cite{jones_characterising_2020}. One specific application of a digital twin is in the detection of anomalies on an aircraft electrical generator, as studied by Boulfani \textit{et al.} \cite{boulfani_anomaly_2020}. The study compares several ML algorithms for early failure detection. The algorithms are trained on anomaly-free test data and are evaluated by using data from flights with anomalies. Other applications of the digital twin include modeling CNC machine tools to predict and diagnose failure \cite{luo_digital_2019}, assisting with product design \cite{tharma_approach_2018}, as well as enhancing privacy in smart cars \cite{damjanovic-behrendt_digital_2018}.

Reservoir computing is an ML algorithm that is particularly well suited for learning dynamical systems on small sets of training data \cite{gauthier_learning_2022}. An RC is commonly trained on time-series data from sensors integrated with the system to predict system dynamics \cite{tanaka_recent_2019}. A traditional RC uses an input layer, a randomly connected artificial recurrent neural network (RNN), and an output layer to generate a prediction. The input-layer and reservoir weights are fixed. Only the output layer is trained via regularized least-squares regression. Since only the output layer is trained, the vanishing gradient problem is avoided \cite{bengio_advances_2013, gauthier_next_2021}. An RC typically requires significantly lower training times than long short term memory (LSTM) or gated recurrent unit (GRU) networks of similar performance \cite{vlachas_backpropagation_2020}. Recent research shows that an alternative version of the RC can learn dynamical systems with even less training data and training time, with fewer metaparameters that require optimization. 

The RC used in this work is an NG-RC, which is mathematically equivalent to a traditional RC but is much less complex. The NG-RC requires no random RNN, it instead forecasts by using trained weights, the time-delay states of the time series data, and nonlinear functionals of the time series data. Since the NG-RC has no random RNN to serve as a reservoir, it has less metaparameters for tuning, making the model easier to train. The feature vector size is also smaller for the NG-RC, which allows the NG-RC to be computationally faster at making forecasts \cite{gauthier_learning_2022}. In this study, the NG-RC is applied to create a digital twin of a dynamical system.

\section{Methods}

\subsection{Numerical simulations of a single scroll turbine engine}

The T-MATS simulation tool was used to generate data that was input to the NG-RC algorithm.  T-MATS provides a physics-based model of the single scroll turbine that includes performance maps for the compressor and turbine.  The dynamic simulator integrates the differential equations of the thermodynamic model and uses interpolation of the performance maps to predict behavior over the full operational envelope of the engine.

\begin{figure}[tb]
    \centering
	\includegraphics[width=5in]{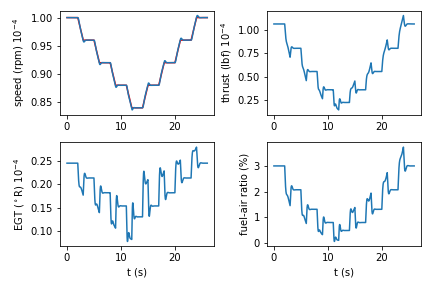}
	\caption{Simulated engine performance. Dynamical evolution of the (top left) requested and actual engine shaft speed, (top right) generated thrust, (bottom right) exhaust gas temperature, and (bottom right) fuel-to-air ratio. The data is sampled every 15 ms, giving 1,734 samples over the full temporal range.}
	\label{fig:TMATS-engine-data}
\end{figure}

A notional flight profile of requested shaft speed was specified. The model proportional-integral speed controller adjusts the fuel-to-air ratio to bring the shaft speed to the requested value in the flight profile. Example data from an engine simulation is shown in Fig. \ref{fig:TMATS-engine-data}. It is seen that the thrust curve has a complex dependence on the shaft speed especially during the graded transition from one shaft speed to the next.  The NG-RC model used a sub-set of this data at different temporal intervals for training the model, and then used the model to predict the thrust at the temporal intervals not used during training.

\subsection{Sensor System}
Jet engine data required to train the ML algorithm was collected by a sensor system that measures several engine parameters. Ambient weather conditions, throttle commands, and engine data were all recorded by the sensor system over several tests of the jet engine. Sensor data was used to train the RC and create a digital twin of the system. The parameters that were measured in this study as well as their sampling rates are shown in Table \ref{tab:1}.

Thrust, intake, exhaust and skin temperatures were all measured with the sensor system, while the rest of the engine parameters were recorded by the factory P100-RX sensor system. The chosen parameters were collected because they are important for modeling the engine with the digital twin. Thrust is an important metric for engine performance, and EGT is one of the most important health monitoring parameters for jet engines \cite{seemann_modeling_2011}.

\begin{figure}[H]
	\centering
	\includegraphics[width=.9\linewidth]{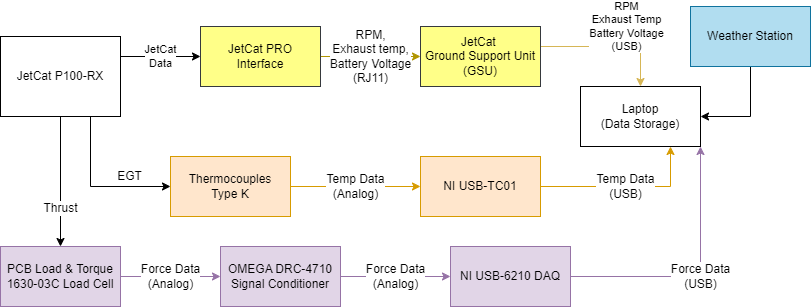}
	\caption{Overview of sensor system}
	\label{fig:p100sensormap}
\end{figure}

The map of the sensor system is also given in Fig. \ref{fig:p100sensormap} to outline how the data was collected. The thrust sensor system consisted of a PCB Load \& Torque Model 1630-03C load cell, an OMEGA DRC-4710 amplifier to condition the small analog signal from the load cell, and an NI USB-6210 data acquisition system to read this analog signal and send the data to a PC. The load cell stops the P100-RX from sliding across linear bearings, so all thrust from the engine was transferred into the load cell. Thrust was sampled at 1000 Hz, temperature data was read at 25 Hz, and all other engine data was read from the P100-RX's ECU at 10 Hz. Engine data from the ECU was received over a USB port with JetCat's built-in serial protocol. Ambient weather conditions were recorded at the beginning of every test from Ohio University's Scalia Laboratory for Atmospheric Analysis. Temperature data was collected by a National Instruments USB-TC01 with type-K thermocouples. Three separate thermocouples recorded the temperature of the jet's skin, air intake, and EGT at 25 Hz. EGT was also collected by a factory installed thermocouple inside of the P100 at 10 Hz. JetCat includes other sensors in the P100-RX to measure engine RPM, pump volts, and battery voltage, which are recorded at 10 Hz as well. All data collected by the sensor system was time-series, meaning it can be used as a direct input to the RC.

\begin{table}[H]
	\caption{Measured Parameters}
	\begin{center}
		\begin{tabular}{|c|c|c|}
			\hline
			\textbf{Parameter}&\textbf{Units}&\textbf{Sampling Rate [S/s]} \\
			\hline
			Thrust&[N]&1000\\
            Intake Temp&[$^\circ$C]&25\\
            EGT&[$^\circ$C]&25\\
            Skin Temp&[$^\circ$C]&25\\
			Engine RPM&[r/min]&10\\
			Engine RPM Setpoint&[r/min]&10\\
			Pump Volts&[V]&10\\
			Battery Voltage&[V]&10\\
            Engine Current&[A]&10\\
			\hline
		\end{tabular}
		\label{tab:1}
	\end{center}
\end{table}

The thrust sensor system was calibrated before data collection for every run of the engine. The procedure used to calibrate the load cell was identical for all the data sets used to train the ML algorithms. Thrust calibration was required to ensure changes in setup, input voltage, and ambient temperature were accounted for in every set of data collection. During calibration of the load cell, several known weights were hung on the load cell to generate voltages. The sensor system recorded thousands of data points for each weight, and these data points were averaged together to find the expected voltage output for the given weight. Six different weights were used to place the load cell in both tension and compression, giving 12 data points. Another reading was made with zero load for a total of 13 data points for each calibration. The data points gathered during the calibration process were used to generate a linear least squares approximation of the thrust sensor system's output. Hysteresis in the load cell was avoided by hanging the weights in an identical randomized pattern for each calibration.

\begin{figure}[H]
    \centering
	\includegraphics[width=3.5in]{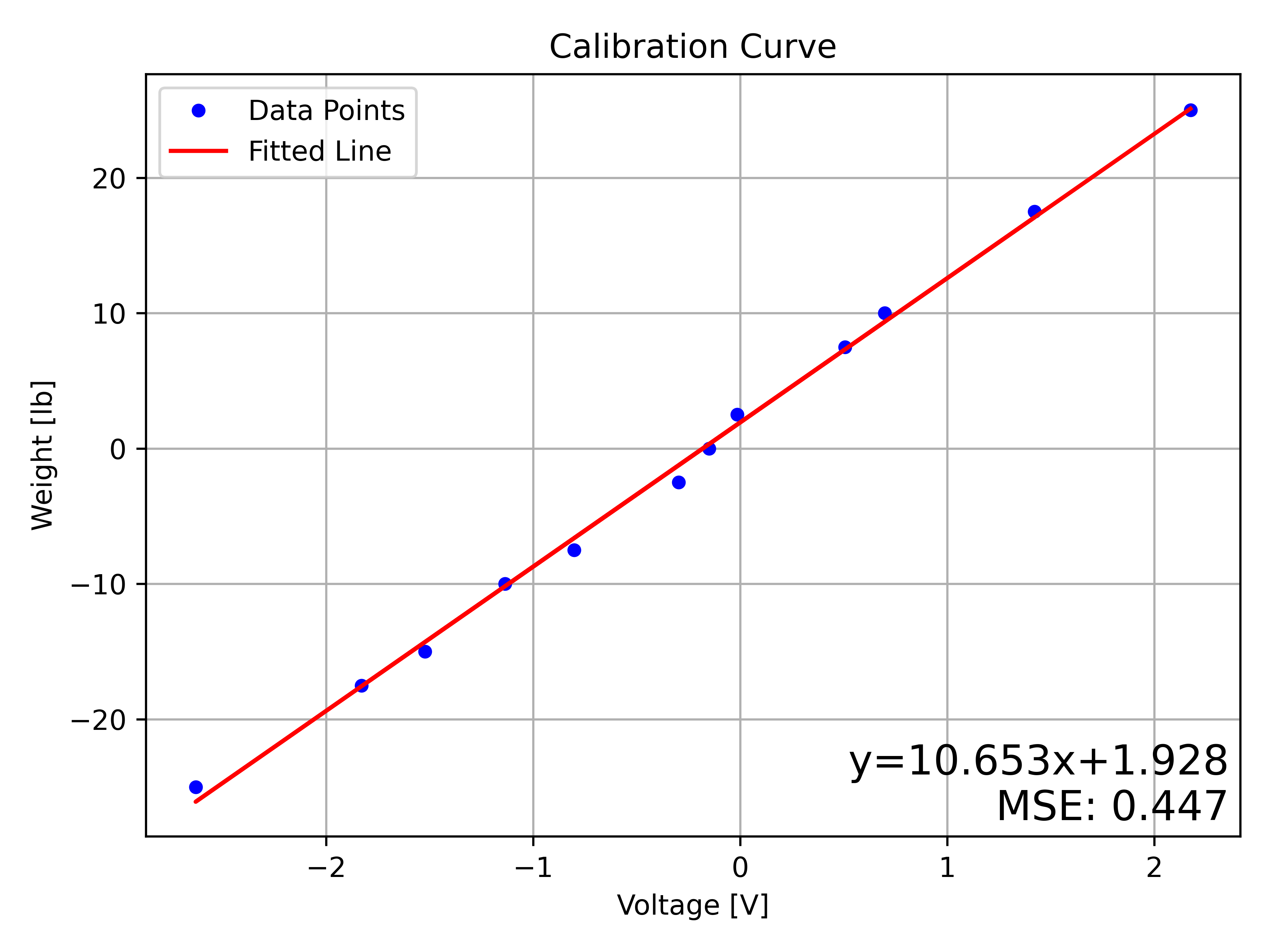}
	\caption{A thrust curve used to calibrate the load cell.}
	\label{fig:thrust curve}
\end{figure}

One example calibration curve creating using this procedure is shown in Fig. \ref{fig:thrust curve}. Each data point from the calibration procedure is shown as well as the line used to fit the data and generate thrust measurements. Mean Squared Error (MSE) was used as an estimate of the quality of fit. The load cell was calibrated using the same procedure before every set of data collection. It is typical to gather calibration data in a manner similar to what is outlined here, and to use a simple line of best fit to determine thrust from voltage \cite{markusic_thrust_2004, desrochers_ground_2001, abdullah_load_2012}.

\subsection{Throttle Control}

Throttle commands were sent to the P100-RX through the THR-AUX pin on the engine's ECU. An Arduino MEGA 2560 microcontroller was used to send a PWM signal to this throttle pin, allowing for simple programs to be written to send the engine identical throttle commands over multiple runs. The microcontroller was programmed to modulate the RPM of the engine according to a user defined throttle curve. Repeatable engine testing and fine control over the engine's RPM set-point was possible with the programmed microcontroller. Repeatable testing was required to evaluate the performance of the NG-RC across different tests with differences in setup and ambient weather conditions.

\subsection{Testing}

\begin{figure}[H]
	\centering
    \includegraphics[width=0.6\linewidth]{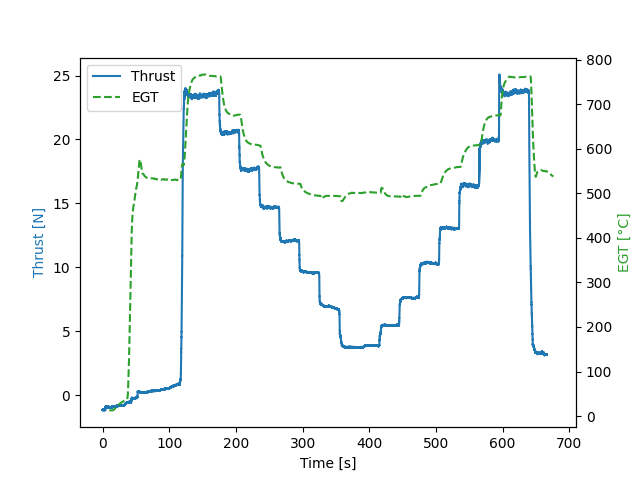}
    \caption{Example data from the sensor system for a descending/ascending test of the P100.}
	\label{fig:example_thrust}
\end{figure}

Several types of engine tests were ran to gather data from the engine under different circumstances. Unit step, ascending step, descending step, and eccentric tests were ran. During the unit step test, the engine was throttled from idle to maximum RPM as fast as possible. The unit step test was ran a total of 15 times. The unit step test makes the fastest, most drastic single change in RPM the engine is capable of performing. Descending and ascending step tests capture the engine making several consistent transitions in RPM. Eccentric tests were also used to observe the engine making many random shifts over its entire RPM range. Random changes assist in detecting any unexpected behavior that might arise by forcing the engine to make many rapid changes in a short time span. Eccentric tests also provide high variance in the training data so that the RC can learn with no hysteresis involved. Example engine data from descending/ascending testing is shown in Fig. \ref{fig:example_thrust}. Correlation between thrust and EGT can be seen, increased thrust leads to a higher EGT. 

\subsection{ML Methods}
The data collected by the sensor system was parsed into validation data and training data to train and evaluate the RC. Data from a single run was split into 10-16 temporal slices. Half of the slices were used to train the NG-RC. The other half were used to evaluate the NG-RC.  For creating an NG-RC model for the simulated engine data, the single data set shown in Fig. \ref{fig:TMATS-engine-data} was used, split into 4 training slices and 5 testing slices.

 For the experimental data collected from the JetCat P100-RX engine, the number of slices varied depending on the run, and were selected to provide a representative sample of the data for both training and evaluation. Next, separate engine runs were used for training and evaluation. Runs were completed on the same day, and the engine was allowed to cool in between runs. Unique flight profiles were used for training and evaluation. Finally, partial data sets from separate runs were combined and used for training and evaluation. For example, the first half of run 1 and the second half of run 2 were used to train the RC, then the second half of run 1 and first half of run 2 were used to evaluate the RC.

The ML model is based on a \textit{feature vector} $\mathbf{o}$, which is an abstract data vector containing the sensor data and representing different physical quantities. At each time step, the feature vector is formed by concatenating a constant, linear features of the measured engine data at the current time step and $k$ past time steps, and nonlinear features that are the unique quadratic monomials of the linear features \cite{gauthier_next_2021}. Using data at the current and past time steps gives short term `memory' to the model, which is needed to accurately model dynamical systems described mathematically by differential equations.

Nonlinear features are required because the engine responds in a nonlinear fashion with respect to the requested shaft speed as can be seen clearly in Fig. \ref{fig:TMATS-engine-data} and \ref{fig:example_thrust}.  Using quadratic monomials is motivated by observing that the thrust depends approximately quadratically on the shaft speed.  For $k>0$, the nonlinear features also contain monomials for quantities at the current time and past times, which gives a measure of the correlation between components.  The model only used measured data; no physical model was required and no compressor or turbine performance maps were needed.

The NG-RC model posits that the desired predicted or inferred values at time step $\mathbf{y}_n$ is related to the feature vector at time step $\mathbf{o}_n$ through the relation
\begin{equation}
\mathbf{y}_n = \mathbf{W}_{out} \mathbf{o}_n. \label{eq:NG-RC}
\end{equation}
For the problem considered here, $\mathbf{y}_n$ is a scalar consisting of the engine thrust data.

During training of the model (finding $\mathbf{W}_{out}$), $N_{train}$ features and associated ground-truth (thrust) data are gathered in blocks $\mathbf{O}$ and $\mathbf{Y}$, respectively.  The weight matrix is then found through the relation
\begin{equation}
\mathbf{W}_{out} = \mathbf{Y} \mathbf{O}^T (\mathbf{O}\mathbf{O}^T+\alpha \mathbf{I})^{-1}, \label{eq:W}
\end{equation}
where $T$ is the matrix transpose operator, $\mathbf{I}$ is the identity matrix, and $\alpha$ is the ridge regression parameter, which regularizes the regression to prevent over fitting of the data.  

During deployment of the model, such as in the testing phase, the weight matrix found from equation ~\ref{eq:W} was used in equation ~\ref{eq:NG-RC} on a step-by-step basis. No physical model of the engine is used to make this prediction, although physical insights guide us to using time-delay linear and quadratic nonlinear components in the feature vector.

For the numerically simulated engine performance data, the linear portion of the feature vector consisted of requested and actual shaft speed, EGT, and fuel-to-air ratio.  The JetCat P100-RX model used the requested and actual shaft speed, the fuel pump voltage, and the turbine skin temperature.  In both cases, the data were converted to dimensionless quantities between 0 and 1.  For the experiments, the thrust data was downsampled to match the sampling rate of the P100-RX's ECU.

The NG-RC has three metaparameters that were adjusted to reduce the prediction error: the lookback or memory parameter $k$, the skip parameter, and the ridge regression parameter $\alpha$. Lookback and skip values from 1-3 were tested, and $10^{-8} \leq \alpha \le 10^{-1}$. 

\section{Results}

\subsection{Creating a ML-based digital twin of a simulated turbine}

\begin{figure}[H]
	\centering
    \includegraphics[width=0.7\linewidth]{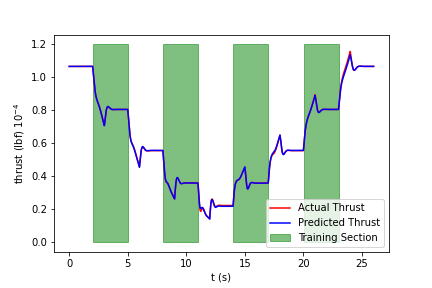}
    \caption{Predicted engine thrust using the optimized NG-RC model.  Temporal evolution of the actual thrust found from simulation of the dynamic T-MATS simulator and predicted thrust from the NG-RC model. The green vertical bars indicate the temporal slices used for training containing 876 temporal points; the rest of the data is used during testing to infer the thrust from the NG-RC model. Metaparameters: $k=1$ and $\alpha = 10^{-5}$.}
	\label{fig:TMATS-RC-comparison}
\end{figure}

Fig. \ref{fig:TMATS-RC-comparison} compares the actual and learned thrust data from the numerically simulated engine after the optimum metaparameters were identified.  Here, $\mathbf{o}_n$ has 1 constant component, 8 linear components, and 36 nonlinear components for a total size of 36 features.  In the white regions, the NG-RC model infers the thrust based on the learned weight matrix $\mathbf{W}_{out}$.  The normalized root-mean-square error is 1.9\% for this data set.  The NG-RC model captures the complex transient dynamics during a step change in the shaft speed as well as the steady-state behavior on each step over the full range of speeds spanning the range from near idle to near full throttle.

This NG-RC used to generate Fig.~\ref{fig:TMATS-RC-comparison} required only 867 data points for training (representing about 100 seconds of engine data), and required a computational time of only 3.8 ms using a commercial-off-the-shelf (COTS) Raspberry Pi 4 running Python, an uncompiled language.  After the RC was trained, was able to generate a new prediction every 9 $\mu$s, or a potential rate of >110 kHz, also on the same compute hardware. 

\subsection{Creating an ML-based digital twin of the JetCat P100-RX turbine}

\begin{figure}[H]
	\centering
    \includegraphics[width=0.7\linewidth]{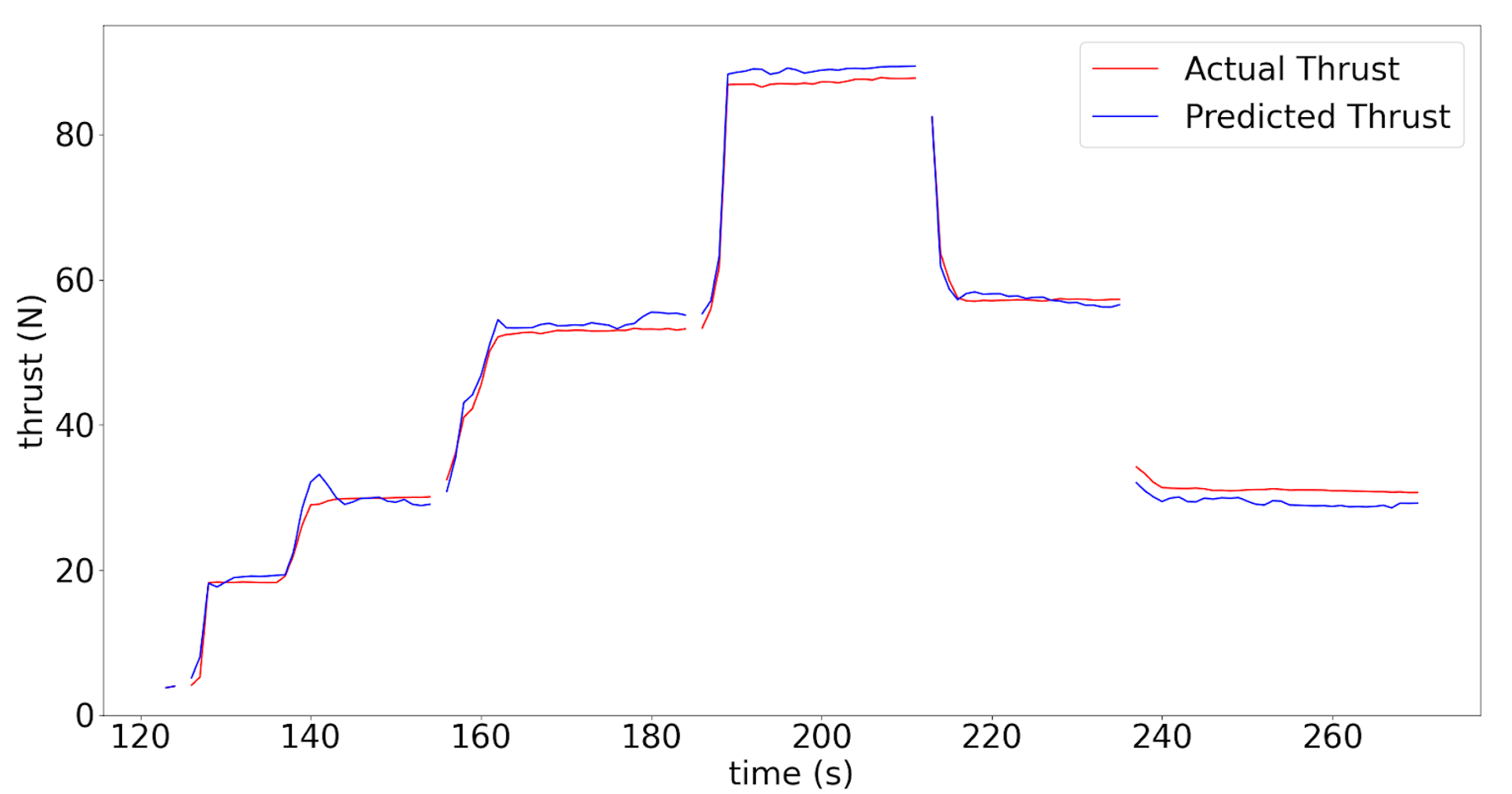}
    \caption{Digital Twin of a JetCat P100-RX created using live engine data.  The RC was able to predict thrust within 1.8\% NRMSE.  Metaparameters: $k=1$ and $\alpha=0.1$.}
	\label{fig:calibrationcurve2}
\end{figure}

Fig. \ref{fig:calibrationcurve2} shows the actual and predicted thrust when the JetCat P100-RX engine was commanded to go through a step-up and step-down sequence.  Shown here is only the testing data; the training data corresponded to the apparent gaps in the data.  The prediction accuracy is similar to the NG-RC model for the simulated data and captures most of the transient features. This NG-RC was trained on 140 data points in only 2.1 ms to predict the output of the engine to within 1.8\% NRMSE.

\begin{figure}[H]
	\centering
    \includegraphics[width=0.7\linewidth]{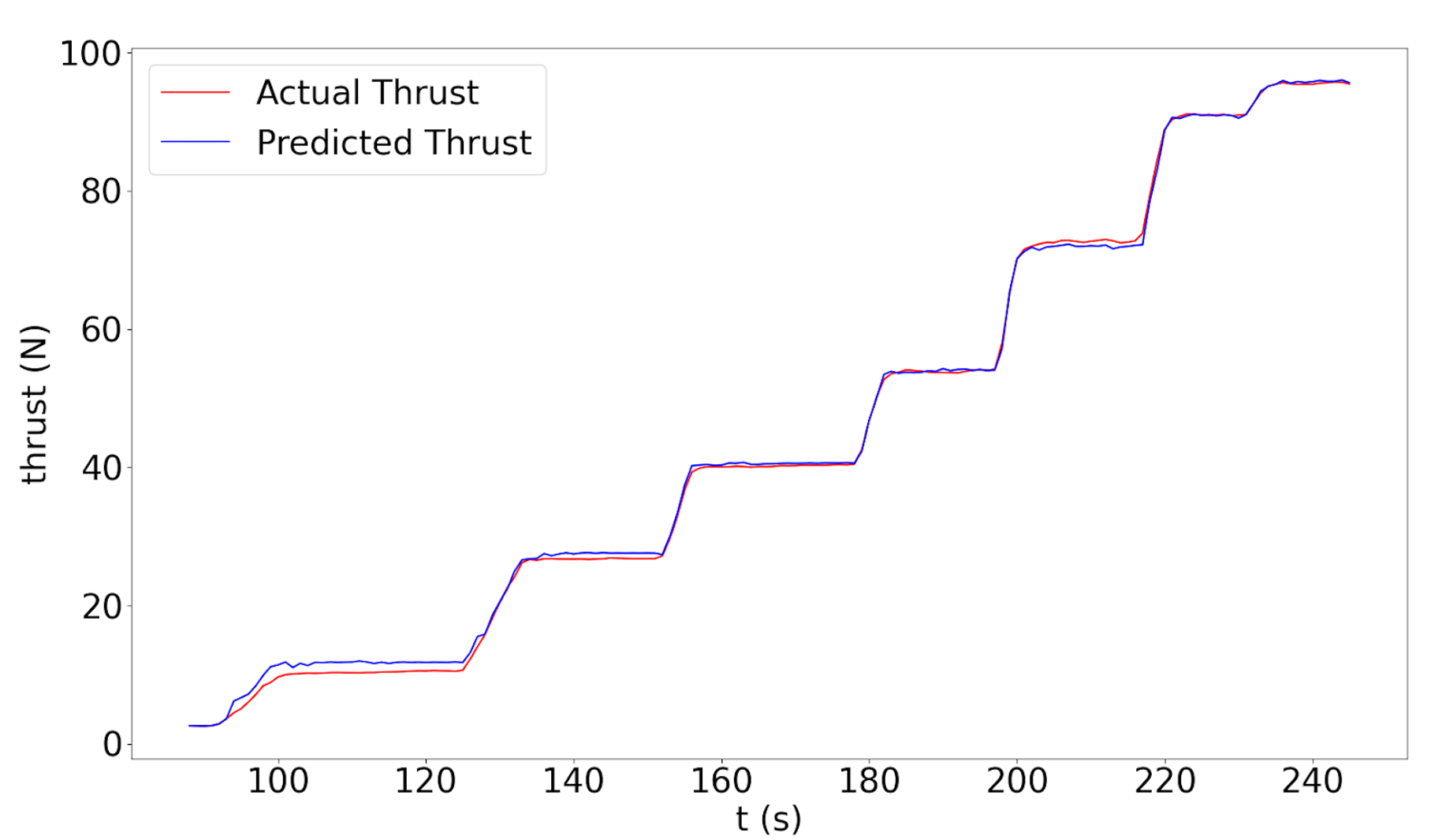}
    \caption{Exploring the generalization of the JetCat Digital Twin.  Results shown are from an RC trained using data from a previous engine run, NRMSE = 0.65\%.}
	\label{fig:calibrationcurve3}
\end{figure}

The generalized accuracy of the JetCat digital twin was explored by investigating the prediction across data sets generated from different engine runs. While the engine itself did not change, there were slight variations in ambient conditions as the time of day changed, the engine was moved and refueled, etc. One example of the Digital Twin’s accuracy when trained on one engine run and tasked to predict the thrust during a subsequent step-up run is shown in Fig. \ref{fig:calibrationcurve3}.

\section{Conclusions}

Engine data was collected from the JetCat P100-RX over many test runs by a sensor system. Thrust, ambient weather conditions, skin temperature, EGT, and intake temperature measured. An NG-RC was trained on this sensor data to create a model of the dynamics of the P100-RX. The NG-RC was trained from small amounts of training data and was used to predict the thrust output of the jet engine. The NG-RC can be trained on 140 data points in only 2.1 ms on a Raspberry Pi 4 single-board computer to predict the output of the engine to within 1.8\% NRMSE. The model was also capable of predicting thrust for subsequent runs based off of training data from previous runs in different weather conditions. The results of this study demonstrated that it was possible to create digital twins of a P100-RX to predict thrust based on EGT, skin temperature and engine RPM via an NG-RC.

The accuracy of the RC-based Digital Twins shown in Figs. \ref{fig:calibrationcurve2} and \ref{fig:calibrationcurve3} are compelling because they were trained on very small data sets. The high accuracy NG-RC demonstrates the small amounts of data required to learn and predict system behavior, which allows for high speed and efficiency.  The small amount of required training data and high efficiency of the NG-RC also shows how the approach can be deployed to low-power edge systems and could be used for controlling the engine to a requested thrust rather than a requested engine speed.

\section*{Acknowledgements}
This material is based upon work supported by the United States Air Force AFRL/RG under Contract No. FA864922P0025.

\bibliography{refs}

\end{document}